\documentclass{article}

\usepackage{arxiv}
\usepackage[utf8]{inputenc}
\usepackage[T1]{fontenc}
\usepackage[hidelinks]{hyperref}
\usepackage{url}
\usepackage{booktabs}
\usepackage{amsfonts}
\usepackage{amsmath}
\usepackage{nicefrac}
\usepackage{microtype}
\usepackage{graphicx}
\usepackage{cleveref}
\usepackage{subcaption}
\usepackage{xcolor}
\usepackage{array}
\usepackage[export]{adjustbox}
\usepackage{authblk}

\definecolor{cBeet}{RGB}{0,0,0}
\definecolor{cOat}{RGB}{80,76,76}
\definecolor{cMeadow}{RGB}{2,81,147}
\definecolor{cRapeseed}{RGB}{152,197,233}
\definecolor{cHop}{RGB}{99,161,199}
\definecolor{cSpelt}{RGB}{219,215,202}
\definecolor{cTriticale}{RGB}{161,172,3}
\definecolor{cBean}{RGB}{228,114,34}
\definecolor{cPea}{RGB}{105,8,89}
\definecolor{cPotato}{RGB}{14,27,95}
\definecolor{cSoybean}{RGB}{4,118,137}
\definecolor{cAsparagus}{RGB}{0,124,49}
\definecolor{cWheat}{RGB}{101,154,29}
\definecolor{cWinterBarley}{RGB}{255,221,0}
\definecolor{cRye}{RGB}{250,186,0}
\definecolor{cSpringBarley}{RGB}{213,75,21}
\definecolor{cMaize}{RGB}{196,72,27}

\definecolor{cCereals}{RGB}{227,39,23}
\definecolor{cWoods}{RGB}{51,196,23}
\definecolor{cForage}{RGB}{248,168,33}
\definecolor{cCorn}{RGB}{251,237,41}
\definecolor{cRice}{RGB}{62,184,225}
\definecolor{cUnkCrop}{RGB}{189,189,189}
\definecolor{cNoAgric}{RGB}{0,0,0}

\title{A Comparative Study of Transformer and Convolutional Models for Crop Segmentation from Satellite Image Time Series}

\author[1]{M. Gatti\thanks{\texttt{mgatti3@uninsubria.it}}}
\author[1]{I. Gallo\thanks{\texttt{ignazio.gallo@uninsubria.it}}}
\author[1]{N. Landro}
\author[1]{C. Loschiavo}
\author[1]{A. U. Rehman.}

\author[2]{M. Boschetti}

\author[3]{R. La Grassa}

\affil[1]{University of Insubria, 21100 Varese (VA), Italy}
\affil[2]{IREA CNR, 20133 Milan (MI), Italy}
\affil[3]{INAF-Astronomical Observatory, 35100 Padua (PD), Italy}

\renewcommand{\shorttitle}{Transformer and Convolutional Models for Crop Segmentation from SITS}

\hypersetup{
pdftitle={A Comparative Study of Transformer and Convolutional Models for Crop Segmentation from Satellite Image Time Series},
pdfsubject={Remote sensing, crop mapping, satellite image time series},
pdfauthor={M. Gatti, I. Gallo, N. Landro, C. Loschiavo, A. U. Rehman, M. Boschetti, R. La Grassa},
pdfkeywords={crop mapping, remote sensing, satellite image time series, transformer models, convolutional models},
}

\date{}

\begin{document}
\maketitle

\begin{abstract}
Crop segmentation from satellite image time series (SITS) is a fundamental task for agricultural monitoring and land-use analysis. While convolutional neural networks (CNNs) have been widely used, transformer-based architectures offer alternative mechanisms for representing spatial and temporal dependencies in multispectral data. This paper presents a comparative study of CNN and transformer-based segmentation models for crop mapping from Sentinel-2 time series, including 3D U-Net, 3D FPN, 3D DeepLabv3, and three transformer architectures: Swin UNETR, TSViT, and VistaFormer, which adopt different strategies for capturing temporal dependencies. Experiments on the Munich and Lombardia datasets show that TSViT achieves the best overall results, slightly surpassing 3D U-Net, which remains a strong CNN baseline. VistaFormer offers the best efficiency, while Swin UNETR performs competitively but is less effective than transformers that explicitly model temporal dynamics. These results highlight that temporal modelling is critical for SITS: TSViT outperforms CNNs and approaches that treat time as an additional spatial dimension, while VistaFormer provides a strong efficiency--performance trade-off.
\end{abstract}


\section{Introduction}

Deep learning has become a widely adopted framework for remote sensing analysis, as highlighted by recent papers on the topic~\cite{yuan2021review,APARNA2022578}. Within this broader context, convolutional neural networks (CNNs) have been particularly successful because of their ability to learn hierarchical spatial representations from multispectral imagery.

In crop mapping, CNN-based models have shown strong performance both on single Sentinel-2 images and on satellite image time series (SITS). For example, CNN approaches have been used successfully for crop segmentation from single-image inputs~\cite{10.1007/978-981-15-6353-9_29,NOWAKOWSKI2021102313}, while temporal approaches based on Sentinel-2 time series have demonstrated the value of explicitly modelling crop evolution over time~\cite{ijgi10070483,russwurm2018multi}. In this setting, widely used segmentation architectures such as U-Net, Feature Pyramid Network (FPN), and DeepLabV3+ have become standard baselines for Sentinel-2 image analysis~\cite{rs12152422unet,rs11020119deeplab,james2021convolutional}.

More recently, transformer-based architectures have emerged as a powerful alternative to CNNs in computer vision. The Vision Transformer (ViT) showed that pure transformer models can achieve strong image classification performance~\cite{DBLP:journals/corr/abs-2010-11929vit}. Transformer-based models were then extended successfully to object detection~\cite{10.1007/978-3-030-58452-8_13obj} and semantic segmentation~\cite{Zheng_2021_CVPRseg}. Among these architectures, the Swin Transformer introduced hierarchical feature representations and shifted-window attention, making transformers more suitable for dense prediction tasks and multi-scale vision problems~\cite{liu2021swin}.

These developments have also influenced remote sensing. Transformer-based models such as SITS-Former~\cite{YUAN2022102651sits}, CTGAN~\cite{9897229ctgan}, UNetFormer~\cite{articleunetformer}, HSI-BERT~\cite{8824217hsibert}, and SpectralFormer~\cite{9627165spectralformer} have reported promising results on remote sensing benchmarks. Additional studies have further confirmed the potential of attention-based models in Earth observation and hyperspectral analysis~\cite{rs13030516,MOHAMMADIMANESH201878}.

In crop mapping, Transformers have also demonstrated strong performance for crop segmentation from single Sentinel-2 images~\cite{wang2022cctnet,NIU2022107297,rs14020359,JAMALI2022103095}. However, crop segmentation from multispectral SITS remains less explored, and it is still unclear how self-attention-based models compare with CNNs when spatial and temporal information must be modeled jointly.

Recent architectures specifically designed for SITS, such as VistaFormer~\cite{macdonald2024vistaformerscalablevisiontransformers} and TSViT~\cite{tarasiou2023vits}, adopt different strategies to model the temporal dimension. TSViT explicitly factorizes temporal and spatial attention, first learning temporal signatures for each spatial location and then modelling spatial relationships. VistaFormer instead uses 3D gated convolutions to downsample spatiotemporal inputs and applies efficient self-attention over the resulting spatial tokens, with the temporal dimension later collapsed in the decoder. In contrast, general-purpose architectures such as Swin UNETR~\cite{hatamizadeh2022swinunetr} treat the input as a 3D volume, processing all dimensions uniformly without explicitly modelling temporal dynamics.

This paper investigates these differences by evaluating CNN and transformer-based models for crop segmentation from Sentinel-2 time series, including an adaptation of Swin UNETR. The implementation of all models used in the experiments is publicly available~\cite{torch_code}.
\section{Architectures}

\begin{figure}[t]
	\begin{center}
    \includegraphics[width=\linewidth]{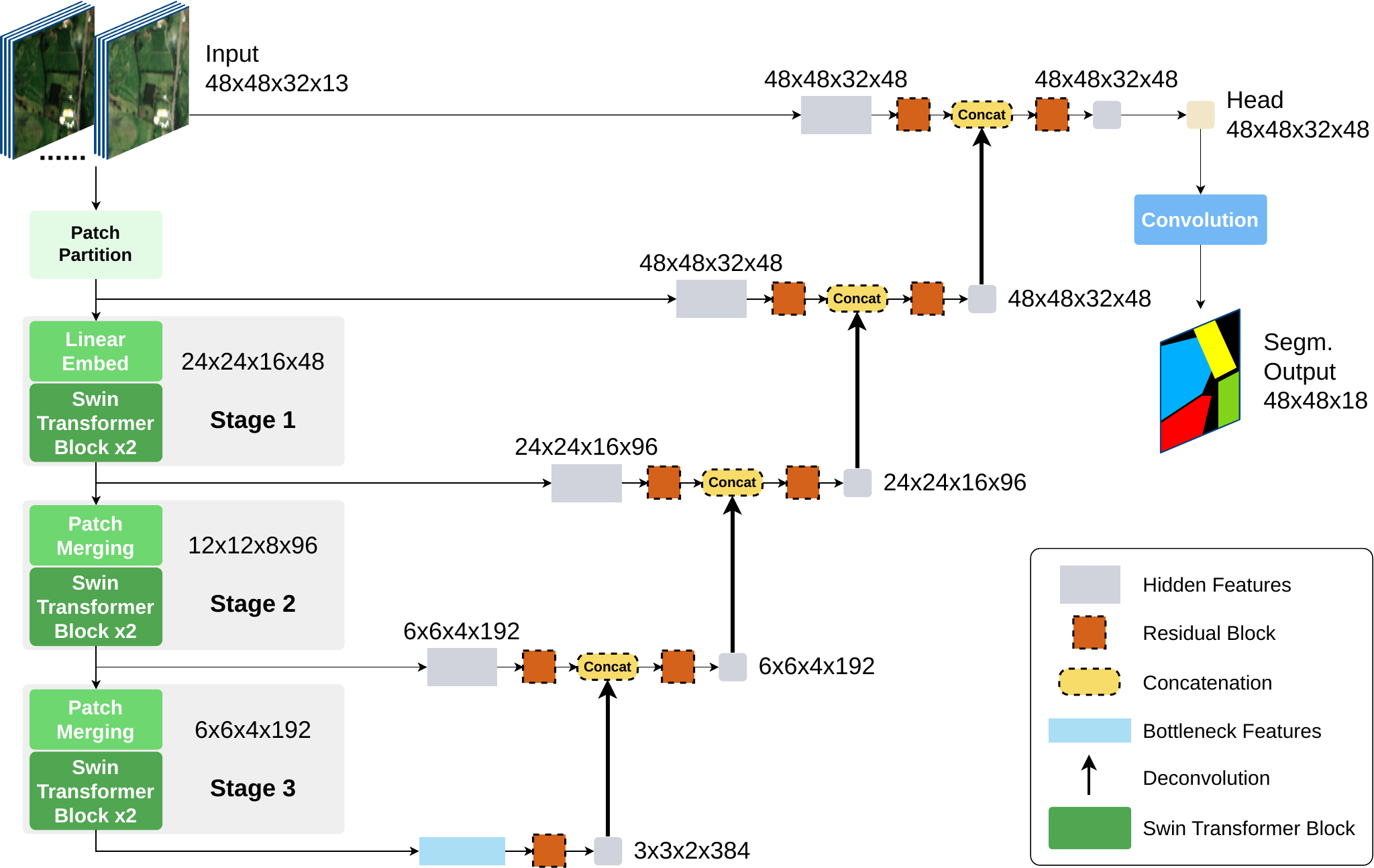}
		\caption{Adapted Swin UNETR architecture for crop segmentation from Sentinel-2 time series. The model receives a multispectral temporal stack as input and produces a dense pixel-wise crop map.}
		\label{fig:proposed}
	\end{center}
\end{figure}

The evaluated models include an adapted Swin UNETR, TSViT, and VistaFormer, which differ in how they model spatial and temporal dependencies, as well as in their computational cost and suitability for dense prediction.

In particular, TSViT explicitly separates temporal and spatial attention. It also encodes temporal information using the acquisition date of each image as positional encoding, enabling the model to capture seasonal patterns and irregular temporal sampling in SITS.

VistaFormer relies on 3D gated convolutions for spatiotemporal downsampling followed by efficient spatial attention, while Swin UNETR treats time as an additional dimension of a 3D volume.

Swin UNETR~\cite{hatamizadeh2022swinunetr}, originally introduced for 3D medical image segmentation, combines a transformer encoder with a U-Net-like decoder, enabling hierarchical feature extraction and localized self-attention through shifted windows. This design addresses limitations of the original Vision Transformer (ViT)~\cite{DBLP:journals/corr/abs-2010-11929vit}, particularly for tasks requiring multi-scale representations. It uses a 3D patch partition followed by a linear embedding layer, where the input volume is divided into non-overlapping patches and processed by a hierarchical Swin Transformer encoder. The model consists of four encoding stages, each with two transformer blocks, followed by a decoder with skip connections.

For Sentinel-2 time series, the temporal axis is treated as the depth dimension, while spectral bands act as input channels, so that each token represents a local spatiotemporal cuboid. This formulation requires several modifications to adapt Swin UNETR to the SITS setting.

First, the input is defined as a tensor of size $32 \times 13 \times 48 \times 48$, corresponding to 32 temporal acquisitions, 13 spectral bands, and spatial dimensions of $48 \times 48$ pixels.

Second, the encoder depth is reduced by removing the fourth stage, as the input resolution does not support further downsampling without excessive loss of spatial detail. The resulting architecture includes six transformer blocks.

Third, the output layer is adapted to produce dense predictions of size $C \times 48 \times 48$, where $C$ is the number of crop classes (e.g., $C=18$ for the Munich dataset).

The final architecture is illustrated in \Cref{fig:proposed}. The number of input channels and output classes can be adapted to different datasets without structural changes. For higher spatial resolutions, additional encoding stages can be introduced to preserve multi-scale features.
\section{Datasets}

Two public datasets based on Sentinel-2 SITS are used for the experiments. Both datasets consist of agricultural areas and provide pixel-wise crop annotations.

\paragraph{Munich dataset} The Munich dataset~\cite{russwurm2018multi} consists of $48 \times 48$ pixel patches with 13 spectral bands. Each pixel is labeled with one of 18 crop classes. Patches are extracted from a larger Sentinel-2 image covering an area of approximately $102 \,\text{km} \times 42 \,\text{km}$ located north of Munich (Germany). Each tile corresponds to a ground area of approximately $480\,\text{m} \times 480\,\text{m}$ (about $2.3 \times 10^{5}\,\text{m}^2$).

The dataset is officially split into 5000 training patches (approximately 60\%), 1700 validation patches (approximately 20\%), and 1700 test patches (approximately 20\%), with the split designed to reduce spatial autocorrelation across subsets. Representative samples are shown in \Cref{fig:munich_samples}.

\newcommand{\legendline}[2]{%
\noindent
\begin{tabular}{@{}c@{\hspace{0.5em}}>{\raggedright\arraybackslash}m{\dimexpr\linewidth-1.5em-0.5em\relax}@{}}
\raisebox{-0.5\height}{\color{#1}\rule{1.2em}{1.2em}} &
\raisebox{-0.5\height}{#2}
\end{tabular}\par\vspace{0.2em}
}

\newsavebox{\munichheaderbox}

\begin{figure*}[t]
\centering

\newlength{\colw}
\newlength{\colsep}
\newlength{\gridw}
\newlength{\legendw}
\newlength{\sidegap}
\newlength{\headergap}

\setlength{\colw}{0.095\textwidth}
\setlength{\colsep}{0.008\textwidth}
\setlength{\gridw}{\dimexpr 5\colw + 4\colsep\relax}
\setlength{\legendw}{0.28\textwidth}
\setlength{\sidegap}{0.03\textwidth}

\setlength{\headergap}{0.1em}

\sbox{\munichheaderbox}{%
  \small
  \makebox[\gridw][c]{%
    \makebox[\colw][c]{$t_1$}\hspace{\colsep}%
    \makebox[\colw][c]{$t_8$}\hspace{\colsep}%
    \makebox[\colw][c]{$t_{16}$}\hspace{\colsep}%
    \makebox[\colw][c]{$t_{32}$}\hspace{\colsep}%
    \makebox[\colw][c]{$y$}%
  }%
}

\begin{minipage}[t]{\gridw}
\centering

\usebox{\munichheaderbox}

\vspace{\headergap}

% -------- Row 1 --------
\makebox[\gridw][c]{%
\includegraphics[width=\colw,frame]{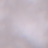}\hspace{\colsep}%
\includegraphics[width=\colw,frame]{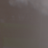}\hspace{\colsep}%
\includegraphics[width=\colw,frame]{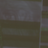}\hspace{\colsep}%
\includegraphics[width=\colw,frame]{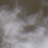}\hspace{\colsep}%
\includegraphics[width=\colw,frame]{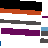}%
}

\vspace{\colsep}

% -------- Row 2 --------
\makebox[\gridw][c]{%
\includegraphics[width=\colw,frame]{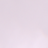}\hspace{\colsep}%
\includegraphics[width=\colw,frame]{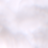}\hspace{\colsep}%
\includegraphics[width=\colw,frame]{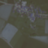}\hspace{\colsep}%
\includegraphics[width=\colw,frame]{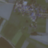}\hspace{\colsep}%
\includegraphics[width=\colw,frame]{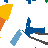}%
}

\vspace{\colsep}

% -------- Row 3 --------
\makebox[\gridw][c]{%
\includegraphics[width=\colw,frame]{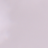}\hspace{\colsep}%
\includegraphics[width=\colw,frame]{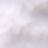}\hspace{\colsep}%
\includegraphics[width=\colw,frame]{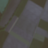}\hspace{\colsep}%
\includegraphics[width=\colw,frame]{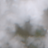}\hspace{\colsep}%
\includegraphics[width=\colw,frame]{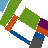}%
}

\end{minipage}%
\hspace{\sidegap}%
\begin{minipage}[t]{\legendw}
\footnotesize

% align legend top with the top of the first image row
\vspace*{\dimexpr\ht\munichheaderbox+\dp\munichheaderbox+\headergap\relax}

\begin{minipage}[t]{\linewidth}
\begin{minipage}[t]{0.48\linewidth}
\raggedright
\legendline{cAsparagus}{Asparagus}
\legendline{cSpringBarley}{Barley (Su.)}
\legendline{cWinterBarley}{Barley (W.)}
\legendline{cBean}{Bean}
\legendline{cHop}{Hop}
\legendline{cMaize}{Maize}
\legendline{cMeadow}{Meadow}
\legendline{cOat}{Oat (Su.)}
\legendline{cPea}{Pea}
\end{minipage}%
\hfill
\begin{minipage}[t]{0.48\linewidth}
\raggedright
\legendline{cPotato}{Potato}
\legendline{cRapeseed}{Rapeseed}
\legendline{cRye}{Rye (W.)}
\legendline{cSoybean}{Soybean}
\legendline{cSpelt}{Spelt (W.)}
\legendline{cBeet}{Sugar beet}
\legendline{cTriticale}{Triticale (W.)}
\legendline{cWheat}{Wheat (W.)}
\end{minipage}
\end{minipage}

\end{minipage}

\caption{Representative samples from the Munich dataset. Each row shows one SITS sample. Columns correspond to selected temporal acquisitions and the corresponding ground-truth segmentation mask. Su. and W. denote summer and winter, respectively.}
\label{fig:munich_samples}
\end{figure*}

\paragraph{Lombardia dataset} The Lombardia dataset~\cite{GALLO2023335} also consists of $48 \times 48$ pixel patches, but includes 9 spectral bands and 7 crop classes. The data are collected from three different geographic areas, referred to as Lombardia1, Lombardia2, and Lombardia3.

\begin{figure*}[t]
\centering

\setlength{\colw}{0.095\textwidth}
\setlength{\colsep}{0.008\textwidth}
\setlength{\gridw}{\dimexpr 5\colw + 4\colsep\relax}
\setlength{\legendw}{0.12\textwidth}
\setlength{\sidegap}{0.03\textwidth}

% gap between column headers and first image row
\setlength{\headergap}{0.1em}

% Save the header row in a box so we can reuse its exact height
\sbox{\munichheaderbox}{%
  \small
  \makebox[\gridw][c]{%
    \makebox[\colw][c]{$t_1$}\hspace{\colsep}%
    \makebox[\colw][c]{$t_8$}\hspace{\colsep}%
    \makebox[\colw][c]{$t_{16}$}\hspace{\colsep}%
    \makebox[\colw][c]{$t_{32}$}\hspace{\colsep}%
    \makebox[\colw][c]{$y$}%
  }%
}

\makebox[\textwidth][c]{%
\begin{minipage}[t]{\gridw}
\centering

\usebox{\munichheaderbox}

\vspace{\headergap}

% -------- Row 1 --------
\makebox[\gridw][c]{%
\includegraphics[width=\colw,frame]{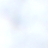}\hspace{\colsep}%
\includegraphics[width=\colw,frame]{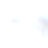}\hspace{\colsep}%
\includegraphics[width=\colw,frame]{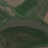}\hspace{\colsep}%
\includegraphics[width=\colw,frame]{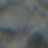}\hspace{\colsep}%
\includegraphics[width=\colw,frame]{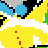}%
}

\vspace{\colsep}

% -------- Row 2 --------
\makebox[\gridw][c]{%
\includegraphics[width=\colw,frame]{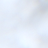}\hspace{\colsep}%
\includegraphics[width=\colw,frame]{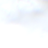}\hspace{\colsep}%
\includegraphics[width=\colw,frame]{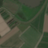}\hspace{\colsep}%
\includegraphics[width=\colw,frame]{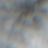}\hspace{\colsep}%
\includegraphics[width=\colw,frame]{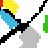}%
}

\vspace{\colsep}

% -------- Row 3 --------
\makebox[\gridw][c]{%
\includegraphics[width=\colw,frame]{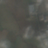}\hspace{\colsep}%
\includegraphics[width=\colw,frame]{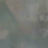}\hspace{\colsep}%
\includegraphics[width=\colw,frame]{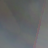}\hspace{\colsep}%
\includegraphics[width=\colw,frame]{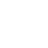}\hspace{\colsep}%
\includegraphics[width=\colw,frame]{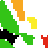}%
}

\end{minipage}%
\hspace{\sidegap}%
\begin{minipage}[t]{\legendw}
\footnotesize

% align legend block with the image block start
\vspace*{\dimexpr\ht\munichheaderbox+\dp\munichheaderbox+\headergap\relax}

% center legend inside a box with the same height as the 3 image rows
\parbox[c][\dimexpr 3\colw + 2\colsep\relax][c]{\linewidth}{%
\raggedright
\legendline{cCereals}{Cereals}
\legendline{cCorn}{Corn}
\legendline{cForage}{Forage}
\legendline{cRice}{Rice}
\legendline{cWoods}{Woods}
\legendline{cUnkCrop}{Unknown crop}
\legendline{cNoAgric}{Non-agricultural}
}

\end{minipage}%
}

\caption{Representative samples from the Lombardia dataset.}
\label{fig:lombardia_samples}
\end{figure*}

For each area, time series are available for four consecutive years (2016--2019), introducing additional temporal variability. The dataset is more challenging due to the diversity of regions and potential inconsistencies in ground truth annotations. The same data split and class definitions as in~\cite{GALLO2023335} are used for training, validation, and testing. Representative samples are shown in \Cref{fig:lombardia_samples}.

\section{Experiments and Discussion}

All models are trained using stochastic gradient descent (SGD) with momentum set to 0.9 and a cosine annealing learning rate scheduler. Training is performed for 200 epochs with a batch size of 32, following the setup in~\cite{GALLO2023335}.

Data augmentation is limited to horizontal and vertical flips. Model performance is evaluated using pixel-level overall accuracy (OA), Cohen's kappa, mean intersection over union (mIoU), and weighted F1-score (wF1), consistent with~\cite{GALLO2023335}. Class-wise precision, recall, and F1-score are computed independently for each crop class, while wF1 is computed as the class-support-weighted average over all classes.

\begin{table}[b]
\centering
\caption{Comparison of segmentation models on the Munich test set. TSViT achieves the highest overall performance, while VistaFormer offers an extremely efficient alternative with minimal computational cost.}
\label{tab:comparison_munich}
\setlength{\tabcolsep}{6pt}
\begin{tabular}{lcccccc}
\toprule
\textbf{Model} & \textbf{Params (M)} & \textbf{FLOPs (G)} & \textbf{OA} & \textbf{Kappa} & \textbf{mIoU} & \textbf{wF1} \\
\midrule
3D DeepLabv3 & 74.95 & 28.07  & 83.43\% & 79.27\% & 52.60\% & 82.85\% \\
3D FPN       & 45.69 & 26.15  & 92.09\% & 90.14\% & 73.02\% & 91.70\% \\
3D U-Net     & 112.18 & 222.74 & 94.39\% & 93.04\% & 81.25\% & 94.28\% \\
Swin UNETR   & 4.50 & 15.67 & 92.51\% & 90.69\% & 75.03\% & 92.36\% \\
TSViT        & 3.26 & 74.02 & \textbf{94.50\%} & \textbf{93.17\%} & \textbf{83.01\%} & \textbf{94.47\%} \\
VistaFormer  & \textbf{0.57} & \textbf{1.97} & 93.60\% & 92.04\% & 79.63\% & 93.50\% \\
Rußwurm et al.~\cite{russwurm2018multi} & -- & -- & 89.60\% & 87.00\% & -- & -- \\
\bottomrule
\end{tabular}
\end{table}

\begin{table}[t]
\centering
\caption{Class-wise performance of TSViT on the Munich dataset. Precision (P), recall (R), and F1-score are reported for validation and test splits together with the number of reference pixels for each class. Su. and W. denote summer and winter, respectively.}
\label{tab:acc_munich}
\setlength{\tabcolsep}{6pt}
\begin{tabular}{lcccccccc}
\toprule
& \multicolumn{4}{c}{\textbf{Validation}} & \multicolumn{4}{c}{\textbf{Test}} \\
\cmidrule(lr){2-5} \cmidrule(lr){6-9}
\textbf{Class} & P & R & F1 & Pixels & P & R & F1 & Pixels \\
\midrule
Asparagus        & 0.68 & 0.92 & 0.79 & 835    & 0.94 & 0.95 & 0.94 & 13867 \\
Barley (Su.)     & 0.93 & 0.93 & 0.93 & 51584  & 0.89 & 0.93 & 0.91 & 52303 \\
Barley (W.)      & 0.96 & 0.96 & 0.96 & 213605 & 0.95 & 0.97 & 0.96 & 170407 \\
Bean             & 0.94 & 0.87 & 0.90 & 15207  & 0.96 & 0.88 & 0.92 & 18944 \\
Hop              & 0.96 & 0.97 & 0.96 & 71411  & 0.96 & 0.97 & 0.96 & 38758 \\
Maize            & 0.97 & 0.98 & 0.97 & 712544 & 0.97 & 0.98 & 0.97 & 603401 \\
Meadow           & 0.91 & 0.88 & 0.90 & 149003 & 0.94 & 0.89 & 0.91 & 166532 \\
Oat (Su.)        & 0.84 & 0.84 & 0.84 & 22765  & 0.86 & 0.80 & 0.83 & 29926 \\
Pea              & 0.83 & 0.89 & 0.86 & 9008   & 0.82 & 0.96 & 0.88 & 5532 \\
Potato           & 0.94 & 0.93 & 0.94 & 74040  & 0.95 & 0.94 & 0.94 & 84205 \\
Rapeseed         & 0.97 & 0.97 & 0.97 & 104936 & 0.98 & 0.98 & 0.98 & 91428 \\
Rye (W.)         & 0.82 & 0.76 & 0.79 & 15314  & 0.82 & 0.78 & 0.80 & 24750 \\
Soybean          & 0.95 & 0.94 & 0.94 & 11944  & 0.96 & 0.92 & 0.94 & 14122 \\
Spelt (W.)       & 0.74 & 0.75 & 0.75 & 17006  & 0.67 & 0.81 & 0.73 & 23281 \\
Sugar beet       & 0.90 & 0.94 & 0.92 & 18783  & 0.97 & 0.98 & 0.97 & 32775 \\
Triticale (W.)   & 0.75 & 0.63 & 0.69 & 33670  & 0.76 & 0.66 & 0.71 & 40552 \\
Wheat (W.)       & 0.96 & 0.96 & 0.96 & 581688 & 0.95 & 0.96 & 0.96 & 529911 \\
\midrule
\textbf{Weighted avg.} & 0.95 & 0.95 & 0.95 &  & 0.95 & 0.94 & 0.94 &  \\
\midrule
\textbf{Overall Accuracy} & \multicolumn{4}{c}{95.04\%} & \multicolumn{4}{c}{94.50\%} \\
\textbf{Cohen's kappa}   & \multicolumn{4}{c}{93.69\%} & \multicolumn{4}{c}{93.17\%} \\
\bottomrule
\end{tabular}
\end{table}

The quantitative comparison on the Munich dataset is reported in \Cref{tab:comparison_munich}. TSViT obtains the best overall performance, slightly surpassing 3D U-Net, which remains the strongest CNN baseline. Swin UNETR also performs well with relatively few parameters, while VistaFormer is the most efficient model and still achieves the best efficiency--accuracy trade-off. The detailed class-wise performance of TSViT is reported in \Cref{tab:acc_munich}.

\begin{figure}[b]
    \centering

    % Group titles
    \makebox[0.44\columnwidth][c]{\textbf{Accurate prediction}}
    \hfill
    \makebox[0.44\columnwidth][c]{\textbf{Challenging prediction}}

    \vspace{0.4em}

    % Accurate prediction
    \begin{minipage}[t]{0.47\columnwidth}
        \centering
        \begin{subfigure}[b]{0.3\linewidth}
            \centering
            \includegraphics[width=\linewidth,frame]{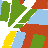}
            \caption{Prediction}
        \end{subfigure}
        \hfill
        \begin{subfigure}[b]{0.3\linewidth}
            \centering
            \includegraphics[width=\linewidth,frame]{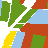}
            \caption{Ground truth}
        \end{subfigure}
        \hfill
        \begin{subfigure}[b]{0.3\linewidth}
            \centering
            \includegraphics[width=\linewidth,frame]{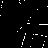}
            \caption{Difference}
        \end{subfigure}
    \end{minipage}
    \hfill
    % Challenging prediction
    \begin{minipage}[t]{0.47\columnwidth}
        \centering
        \begin{subfigure}[b]{0.3\linewidth}
            \centering
            \includegraphics[width=\linewidth,frame]{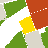}
            \caption{Prediction}
        \end{subfigure}
        \hfill
        \begin{subfigure}[b]{0.3\linewidth}
            \centering
            \includegraphics[width=\linewidth,frame]{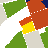}
            \caption{Ground truth}
        \end{subfigure}
        \hfill
        \begin{subfigure}[b]{0.3\linewidth}
            \centering
            \includegraphics[width=\linewidth,frame]{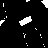}
            \caption{Difference}
        \end{subfigure}
    \end{minipage}

    \caption{Qualitative examples on the Munich dataset. The left group shows an accurate prediction, where the predicted segmentation closely matches the reference annotation, with only minor discrepancies near parcel boundaries. The right group shows a challenging prediction, where misclassifications affect larger contiguous regions, indicating confusion between crop classes with similar spectral or temporal signatures.}
    \label{fig:munich_predictions}
\end{figure}

Qualitative examples are shown in \Cref{fig:munich_predictions}. Accurate predictions are typically observed within homogeneous parcels, where the model correctly captures crop boundaries and class labels. Most errors are concentrated along field borders, where transitions between crop types are more ambiguous. In more challenging cases, misclassifications affect larger contiguous regions, suggesting confusion between crops with similar spectral or temporal signatures.

\setlength{\fboxrule}{0.4pt}

The results on the Lombardia dataset, along with model comparisons, are reported in \Cref{tab:comparison_lombardia}. TSViT achieves the highest OA, Kappa, and wF1, while 3D U-Net attains the highest mIoU, highlighting a slight trade-off between pixel-wise accuracy and region overlap. Overall performance is consistently lower across all models, indicating that this dataset represents a more challenging scenario.

\begin{table}[t]
\centering
\caption{Comparison of segmentation models on the Lombardia test set.}
\label{tab:comparison_lombardia}
\setlength{\tabcolsep}{6pt}
\begin{tabular}{lcccccc}
\toprule
\textbf{Model} & \textbf{Params (M)} & \textbf{FLOPs (G)} & \textbf{OA} & \textbf{Kappa} & \textbf{mIoU} & \textbf{wF1} \\
\midrule
3D DeepLabv3 & 74.86 & 26.45 & 68.07\% & 61.06\% & 50.12\% & 67.17\% \\
3D FPN       & 45.67 & 23.76 & 72.37\% & 66.34\% & 55.17\% & 71.41\% \\
3D U-Net     & 112.18 & 221.95 & 74.46\% & 68.98\% & \textbf{58.50\%} & 74.05\% \\
Swin UNETR   & 4.50 & 15.22 & 71.61\% & 65.40\% & 54.43\% & 71.44\% \\
TSViT        & 3.25 & 53.17 & \textbf{74.74\%} & \textbf{69.26\%} & 57.77\% & \textbf{74.12\%} \\
VistaFormer  & \textbf{0.57} & \textbf{1.93} & 74.24\% & 68.75\% & 57.31\% & 73.53\% \\
\bottomrule
\end{tabular}
\end{table}

\begin{figure}[h]
    \centering

    % Group titles
    \makebox[0.44\columnwidth][c]{\textbf{Accurate prediction}}
    \hfill
    \makebox[0.44\columnwidth][c]{\textbf{Challenging prediction}}

    \vspace{0.4em}

    % Accurate prediction
    \begin{minipage}[t]{0.47\columnwidth}
        \centering
        \begin{subfigure}[b]{0.3\linewidth}
            \centering
            \includegraphics[width=\linewidth,frame]{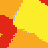}
            \caption{Prediction}
        \end{subfigure}
        \hfill
        \begin{subfigure}[b]{0.3\linewidth}
            \centering
            \includegraphics[width=\linewidth,frame]{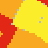}
            \caption{Ground truth}
        \end{subfigure}
        \hfill
        \begin{subfigure}[b]{0.3\linewidth}
            \centering
            \includegraphics[width=\linewidth,frame]{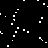}
            \caption{Difference}
        \end{subfigure}
    \end{minipage}
    \hfill
    % Challenging prediction
    \begin{minipage}[t]{0.47\columnwidth}
        \centering
        \begin{subfigure}[b]{0.3\linewidth}
            \centering
            \includegraphics[width=\linewidth,frame]{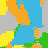}
            \caption{Prediction}
        \end{subfigure}
        \hfill
        \begin{subfigure}[b]{0.3\linewidth}
            \centering
            \includegraphics[width=\linewidth,frame]{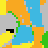}
            \caption{Ground truth}
        \end{subfigure}
        \hfill
        \begin{subfigure}[b]{0.3\linewidth}
            \centering
            \includegraphics[width=\linewidth,frame]{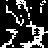}
            \caption{Difference}
        \end{subfigure}
    \end{minipage}

    \caption{Qualitative examples on the Lombardia dataset. The left group shows an accurate prediction, while the right group shows a challenging prediction with more evident spatially structured errors.}
    \label{fig:lombardia_predictions}
\end{figure}

Qualitative examples in \Cref{fig:lombardia_predictions} show that, although the general structure of agricultural parcels is preserved, prediction errors are more frequent and exhibit clear spatial patterns. The higher performance observed on the Munich dataset suggests that the evaluated models benefit from the more consistent spatial and temporal structure of this dataset. In contrast, the lower scores on the Lombardia dataset may be attributed to increased intra-class variability and potential inconsistencies in the ground truth annotations, possibly caused by inaccurate crop declarations.

Among the compared models, 3D DeepLabv3 performs worst, likely due to atrous convolutions~\cite{DBLP:journals/corr/ChenPSA17}, which are less effective on small input patches ($48 \times 48$). 3D U-Net remains a strong baseline, confirming the effectiveness of convolutional architectures for dense prediction. However, transformer-based models, particularly TSViT, obtain the best overall scores in these experiments, suggesting that explicitly modelling temporal dynamics can be beneficial for crop segmentation from SITS. In contrast, Swin UNETR, which treats time as an additional spatial dimension, remains competitive but does not fully exploit temporal structure.
This suggests that the choice of temporal modelling strategy plays a critical role, especially in more challenging and heterogeneous scenarios.
\section{Conclusions}

This paper presented a comparative evaluation of transformer-based architectures for crop segmentation from SITS. The analysis included an adapted Swin UNETR, TSViT, and VistaFormer, alongside established 3D CNN baselines.

Across both datasets, TSViT consistently outperforms the other evaluated models, including both CNN-based and alternative transformer architectures. VistaFormer provides the most efficient solution, while maintaining strong accuracy with significantly lower computational cost, highlighting the benefits of lightweight transformer designs. Swin UNETR also performs well, confirming that treating SITS as 3D volumes is a viable approach, but is less effective than methods that explicitly model temporal information.

Overall, the results highlight the importance of temporal modelling for accurate crop segmentation from SITS. Architectures that explicitly capture temporal dependencies, such as TSViT, provide a clear advantage over both convolutional models and transformers that treat time as an additional spatial dimension, underscoring the central role of temporal representations in SITS analysis. These findings suggest that effectively modelling temporal dynamics is not only beneficial but essential for fully exploiting the information contained in satellite image time series, particularly in scenarios characterized by high intra-class variability and complex seasonal patterns.

Future work will focus on extending this analysis to higher-resolution imagery, more diverse geographic regions, and additional geospatial tasks, as well as further investigating efficient transformer designs for spatiotemporal modelling.

\bibliographystyle{unsrt}  
\bibliography{references}  

\end{document}